%% file: gddl_2026.tex
\documentclass{article}

\PassOptionsToPackage{numbers, compress}{natbib}

\usepackage[dblblindworkshop, final]{neurips_2026}
\workshoptitle{Geometric Distributional Deep Learning: Bridging Optimal Transport, Learning and Structured Data}

\usepackage[utf8]{inputenc} 
\usepackage[T1]{fontenc}    
\usepackage{hyperref}       
\usepackage{url}            
\usepackage{booktabs}       
\usepackage{amsfonts}       
\usepackage{nicefrac}       
\usepackage{microtype}      
\usepackage{xcolor}         
\usepackage{amsmath}        
\usepackage{array}          
\usepackage{float}          

\newcommand{\ptsub}[1]{\text{\scshape #1}}
\newcommand{\PT}[1]{\mathrm{PT}_{\ptsub{#1}}}
\newcommand{\pmx}[1]{{\scriptsize$\pm$#1}}
\newcommand{\rawfeat}{\textsc{RawFeat}}
\newcommand{\dfs}[2]{DFS$_{#1,#2}$}

\title{From Synthetic Databases to Relational Representations: The Role of Curricula, DFS, and Depth in Graph-Based Relational Foundation Models}

\author{%
  Mohammad Sadeq Abolhasani \\
  SAP Labs \\
  \texttt{Rastin.Abolhasani@sap.com} \\
  \And
  Viswanath Ganapathy \\
  SAP Labs \\
  \texttt{viswa.ganapathy@sap.com} \\
}

\begin{document}

\maketitle

\begin{abstract}
Prior-data fitted networks (PFNs) transfer from synthetic pretraining to unseen datasets without fine-tuning. We ask whether this recipe carries over to graph-based relational foundation models, and what governs the transfer. We pretrain the message-passing encoder Griffin on synthetic relational databases, keep it frozen, and read it out with a frozen in-context TabICL head. On 16 real-world classification tasks synthetic pretraining helps, and an ordered curriculum from schema-guided to schema-agnostic data beats every single-stage mixture of the same corpora as well as real-data pretraining alone; appending the curriculum after real-data pretraining is best of all. Frozen synthetic pretraining nevertheless turns out to be only a classification phenomenon: on the full regression suite of eight tasks no frozen encoder beats raw target-row features, and curriculum order, while it still matters, barely moves regression. Hop-bounded Deep Feature Synthesis (DFS) aggregates improve both metrics monotonically with hop depth and are the only frozen change to overtake raw features on regression. Deeper 4-hop subgraphs lie dormant when frozen; fine-tuning the encoder into the head unlocks them and yields the best classification and regression scores in the study.
\end{abstract}

\section{Introduction}

Relational databases (RDBs) hold most enterprise data, yet relational foundation models lag text and vision because real RDBs are private, scarce, and heterogeneous. Prior-data fitted networks (PFNs)~\citep{muller2021pfn,hollmann2022tabpfn} sidestep this by pretraining on an unlimited population of synthetic tasks, and RDB-PFN~\citep{wang2026rdbpfn} extends them to full databases with a transformer over inputs linearized by Deep Feature Synthesis (DFS)~\citep{kanter2015dfs}. Graph-based relational models instead encode foreign-key (FK) structure natively and learn aggregation rather than enumerate it~\citep{fey2023rdl,wang2025griffin,zhang2023gfs}. Whether the PFN paradigm transfers to them, and what governs the transfer, is open.

We study this with one pipeline. The message-passing encoder Griffin~\citep{wang2025griffin} is pretrained on PluRel-generated synthetic databases~\citep{abolhasani2026plurel}, kept frozen, and read out by a frozen in-context TabICL head~\citep{qu2025tabicl}, so frozen downstream performance measures the pretrained representation alone. Pretraining follows ordered curricula~\citep{bengio2009curriculum,abolhasani2026curriculum} over a schema-guided and a schema-agnostic synthetic corpus, optionally initialized from real-data pretraining, and evaluation covers 16 real-world classification and eight regression tasks. We ask: (\textbf{RQ1}) Does synthetic pretraining yield a useful frozen classification representation, and does the order of the curriculum matter beyond mixing its corpora? (\textbf{RQ2}) How does synthetic pretraining compare with, and combine with, real-data pretraining? (\textbf{RQ3}) Does the frozen recipe transfer to regression? (\textbf{RQ4}) Does explicit, hop-bounded DFS aggregation complement learned message passing? (\textbf{RQ5}) Does a deeper 4-hop subgraph help a frozen encoder? (\textbf{RQ6}) Does fine-tuning the encoder into the head unlock performance, and how does it interact with depth?

All results are collected in Tables~\ref{tab:master} to~\ref{tab:axes}; the experimental details, per-task values, and the pairwise comparisons behind every claim are in Appendices~\ref{app:details} to~\ref{app:earlystop}.

\section{Pipeline, data, and evaluation}

\paragraph{Encoder, head, and floor.}
Griffin~\citep{wang2025griffin} is a heterogeneous message-passing network over the FK graph. For a target row it samples a temporal subgraph of depth $h$ hops (fanout 20, few-shot fanout 3), applies four relation-aware message-passing layers with reversed FK edges, and emits a 512-dimensional embedding of the target entity. TabICL~\citep{qu2025tabicl} predicts query labels from an in-context set of labeled examples without gradient updates. We keep it frozen (eight-estimator ensemble, context size 1024) as head and evaluation engine, so frozen downstream performance measures the pretrained representation alone. The floor, \rawfeat{}, runs TabICL on raw target-row features with no encoder.

\paragraph{Synthetic corpora and curricula.}
The schema-guided corpus (\textsc{sg}) holds 500 PluRel-generated databases~\citep{abolhasani2026plurel} that share the four-table FK schema of the real SAP sales database \texttt{rel-salt}, each with seven multi-class tasks. The schema-agnostic corpus (\textsc{sa}) holds 9,143 databases from three generator sub-families, each with a fresh schema and binary mask-a-column tasks. $\PT{x}\!\to\!\PT{y}$ denotes an ordered curriculum whose stage $Y$ starts from the best validation checkpoint of stage $X$, and $\mathrm{PT}_{\ptsub{x}\cup\ptsub{y}}$ a single stage on the shuffled mixture. Every stage runs under one early-stopping protocol (Table~\ref{tab:settings}), so ordered and unordered runs differ only in data order. Real-data pretraining uses two task families and is evaluated only on the other two: $\PT{c2o2}$ pretrains on Com2+Othr2 and is evaluated on Com1+Othr1, and $\PT{c1o1}$ does the reverse.

\paragraph{Downstream tasks and metrics.}
Classification uses 16 RelBench and 4DBInfer tasks~\citep{robinson2024relbench,wang2024fourdbinfer} in four families, Com1 (5 tasks), Com2 (4), Othr1 (5), and Othr2 (2), scored by ROC-AUC under frozen in-context prediction. Regression uses eight tasks in the same family split, two in Com1+Othr1 and six in Com2+Othr2 (task lists in Appendix~\ref{app:details}), scored by $-\mathrm{MAE}/\sigma_{\text{task}}$, where $\sigma_{\text{task}}$ is the standard deviation of the target, so higher (closer to 0) is better and a value near $-1.0$ corresponds to predicting the mean. Each real-initialized run is evaluated only on its held-out subset, and regression is averaged per subset because the two subsets are not comparable; only the fully synthetic runs and \rawfeat{} cover all 16 classification and all eight regression tasks.

\paragraph{Evaluation axes.}
We vary four further things. (i) \emph{DFS augmentation.} \dfs{f}{s} appends explicit DFS aggregates~\citep{kanter2015dfs} to the encoder: statistics of related rows computed up to $f$ hops along the sampled fanout neighborhood and up to $s$ hops along the few-shot (labeled-example) neighborhood. We evaluate \dfs{2}{0} and \dfs{2}{2} on top of real-data pretraining. The augmentation is deliberately lightweight and hop-bounded, unlike the full enumeration RDB-PFN linearizes, and complements learned message passing rather than replacing it. (ii) \emph{Depth.} Subgraphs of $h{=}4$ instead of $h{=}2$ hops, with the four message-passing layers unchanged. (iii) \emph{Fine-tuning (FT).} The encoder is trained jointly with the TabICL head on the downstream task; fine-tuned runs are reported separately and never ranked against frozen ones. (iv) \emph{Regression} for every configuration. Classification values of the curriculum configurations are means over five independent end-to-end seeds and are reported with their standard deviations, as are the DFS-augmented classification results; the depth, fine-tuning, and regression results are single runs.

\section{Results}
\label{sec:results}

\begin{table}[t]
  \caption{Frozen in-context performance by task family (context 1024): ROC-AUC (CLF) and $-\mathrm{MAE}/\sigma$ (REG), ranked by Avg CLF. Real-initialized runs are evaluated only on held-out families. Avg REG is over the run's regression subset, A (Com1+Othr1, 2 tasks), B (Com2+Othr2, 6 tasks), or C (all); marks are not comparable. $\pm$: std over five seeds. Bold: best per column (REG per mark).}
  \label{tab:master}
  \centering
  \footnotesize
  \setlength{\tabcolsep}{3.2pt}
  \input{tables/master}
\end{table}

\paragraph{Synthetic pretraining and curriculum order (RQ1).}
Table~\ref{tab:master} ranks every frozen configuration by mean ROC-AUC over the 16 classification tasks, and Table~\ref{tab:matched} restricts each comparison to the task subset on which all compared runs are evaluated. Every pretrained encoder lies above the no-encoder floor \rawfeat{}, and the three purely synthetic recipes form a ladder, $\PT{sg}$ $<$ $\PT{sa}$ $<$ $\PT{sg}\!\to\!\PT{sa}$: synthetic pretraining helps, the larger schema-agnostic corpus helps more, and opening with the schema-guided stage before it helps most. Order rather than mixing drives the curriculum. Table~\ref{tab:curriculum} pairs each ordered curriculum with the single-stage mixture of exactly the same data, and the ordered run wins all three pairs on both metrics, on classification by margins larger than the seed standard deviation in every pair.

\begin{table}[t]
  \caption{Matched task-subset comparison (context 1024); each column compares only runs evaluated on that subset (task count in parentheses). Bold: best per column. On regression, only the DFS-augmented runs on Com1+Othr1 beat \rawfeat{}.}
  \label{tab:matched}
  \centering
  \footnotesize
  \input{tables/matched}
\end{table}

\paragraph{Synthetic versus real pretraining data (RQ2).}
On matched subsets (Table~\ref{tab:realsynth}) the purely synthetic curriculum beats real-data pretraining alone in both directions, and appending the synthetic curriculum after real-data pretraining is best overall: a clear gain over the synthetic curriculum in the C2O2 direction and a tie within one seed standard deviation in the C1O1 direction. Synthetic and real data are therefore complementary rather than substitutes, and the real-to-synthetic continuation is the strongest frozen classification encoder in the study.

\paragraph{Regression does not inherit the classification story (RQ3).}
Table~\ref{tab:axes}(c) ranks the configurations evaluated on all eight regression tasks, and the no-encoder floor \rawfeat{} leads every frozen encoder; the matched subsets of Table~\ref{tab:matched} show the same picture. The curriculum effects survive only in weakened form: ordered still beats unordered in all three pairs and real-to-synthetic still beats real alone in both directions (Tables~\ref{tab:curriculum} and~\ref{tab:realsynth}), but no frozen encoder without DFS reaches the raw-feature floor of its subset. The recipe that lifts classification well clear of raw features (Table~\ref{tab:floorgains}) thus transfers to classification only: a frozen relational representation loses regression signal that the raw row carries. Only two kinds of intervention get past raw features, and only on Com1+Othr1: DFS augmentation of a frozen encoder, clearly for \dfs{2}{2} and marginally for \dfs{2}{0} (Table~\ref{tab:matched}), and fine-tuning (Table~\ref{tab:axes}(b)); on Com2+Othr2 nothing does.

\begin{table}[t]
  \caption{The added axes: (a) frozen levers on real-data pretraining (RQ4, RQ5), each direction on its held-out subset; (b) fine-tuning the encoder into the head (FT, RQ6), with REG on the run's subset (A, B, C as in Table~\ref{tab:master}); (c) regression over all eight tasks (RQ3). DFS helps monotonically with hop depth, frozen depth does not help, and fine-tuning unlocks it.}
  \label{tab:axes}
  \centering
  \footnotesize
  \setlength{\tabcolsep}{3pt}
  \begin{minipage}[t]{0.62\linewidth}
    \centering
    \input{tables/levers}
  \end{minipage}\hfill
  \begin{minipage}[t]{0.36\linewidth}
    \centering
    \input{tables/full8}
  \end{minipage}\\[1pt]
  \begin{minipage}[t]{\linewidth}
    \centering
    \input{tables/finetune}
  \end{minipage}
\end{table}

\paragraph{Hop-bounded DFS complements message passing (RQ4).}
Table~\ref{tab:axes}(a) shows the same ladder in both held-out directions and on both metrics: \dfs{2}{2} $>$ \dfs{2}{0} $>$ no DFS. The second step comes purely from extending the aggregates over the few-shot neighborhood ($s$ from 0 to 2): labeled neighbors carry signal that frozen message passing misses. Regression is where DFS matters most: \dfs{2}{2} lifts regression in both directions by more than any curriculum change does (Tables~\ref{tab:matched} and~\ref{tab:curriculum}), and the DFS-augmented runs are the only frozen configurations to overtake \rawfeat{} on any subset, on Com1+Othr1 (Table~\ref{tab:matched}). On classification, in each direction the DFS-augmented runs rank above real-data pretraining alone and below the real-to-synthetic curriculum; \dfs{2}{2} beats the purely synthetic curriculum in the C2O2 direction and trails it in the C1O1 direction (Table~\ref{tab:matched}), so DFS adds signal but does not replace the curriculum. This supports a design in which learned aggregation does most of the work and a small, hop-bounded explicit complement helps, without RDB-PFN's full enumeration.

\paragraph{Depth is dormant when frozen and unlocked by fine-tuning (RQ5, RQ6).}
A frozen 4-hop encoder is at or below its 2-hop counterpart on both metrics in both directions (Table~\ref{tab:axes}(a)): the deeper context is capacity the frozen pipeline cannot exploit. Fine-tuning changes this (Table~\ref{tab:axes}(b)): it lifts both metrics for every encoder; the gain is smallest for the strongest frozen encoder, so adaptation partly substitutes for pretraining quality, and largest for the 4-hop $\PT{c2o2}$ encoder, which becomes the study's best encoder on both metrics: on its subset it overtakes the best frozen curriculum on classification and is the only configuration ahead of both \rawfeat{} and \dfs{2}{2} on regression (Table~\ref{tab:floorgains}). Depth thus pays off only with adaptation, and the gap it closes is largest for regression.



\section{Limitations and conclusion}

The regression, depth, and fine-tuning results are single runs, so we treat them as trends rather than precise estimates. The \dfs{2}{0} step is the smallest effect we report and is best confirmed with more seeds, while the \dfs{2}{2} gains, the depth gap, and the fine-tuning gains are several times larger. Absolute regression scores stay near the predict-the-mean reference across pipelines, so we read them as relative comparisons between recipes. DFS and $4$-hop sampling were added only to real-data pretraining; combining them with the synthetic curriculum, and fine-tuning DFS-augmented encoders, are natural next steps. A secondary early-stopping ablation (Appendix~\ref{app:earlystop}) is consistent with the main results.

Overall, a frozen, synthetically pretrained message-passing encoder gives a strong classification representation whose quality tracks curriculum order, while the same recipe is less effective for regression, where raw features remain a competitive baseline. Hop-bounded DFS aggregates recover part of that regression gap and improve with hop depth, and fine-tuning the encoder into the in-context head recovers more of it while also making deeper subgraphs useful, yielding the best encoder we observe. We hope these observations are a useful step toward relational foundation models.


\appendix
\raggedbottom

\section{Experimental details}
\label{app:details}
Table~\ref{tab:settings} collects the pipeline, corpus, and protocol settings used throughout, and Table~\ref{tab:tasks} lists every downstream task by family. Context size is 1024 for every evaluation. Real-data pretraining on Com2+Othr2 ($\PT{c2o2}$) is evaluated on Com1+Othr1 (10 classification and 2 regression tasks), and real-data pretraining on Com1+Othr1 ($\PT{c1o1}$) on Com2+Othr2 (6 classification and 6 regression tasks), so no run is evaluated on a family it was pretrained on. Fully synthetic runs and \rawfeat{} are evaluated on all 16 classification and all 8 regression tasks; their regression average over all 8 tasks equals the task-weighted mean of the two subset averages.

\begin{table}[H]
  \caption{Pipeline, data, and protocol settings.}
  \label{tab:settings}
  \centering
  \footnotesize
  \begin{tabular}{@{}>{\raggedright\arraybackslash}p{0.25\linewidth}>{\raggedright\arraybackslash}p{0.71\linewidth}@{}}
  \toprule
  Component & Setting \\
  \midrule
  Encoder & Griffin: heterogeneous message passing over the FK graph with reversed FK edges; four relation-aware layers; 512-dimensional embedding of the target entity. \\
  Subgraph sampling & Temporal subgraph of the target row; depth $h{=}2$ by default and $h{=}4$ for the depth axis; fanout 20; few-shot (labeled-example) fanout 3. \\
  Head & TabICL, frozen: eight-estimator ensemble, context size 1024, in-context prediction with no gradient updates; also the evaluation engine for every configuration. \\
  Floor & \rawfeat{}: TabICL on the raw target-row features with no encoder. \\
  Schema-guided corpus (\textsc{sg}) & 500 PluRel-generated databases sharing the four-table FK schema of \texttt{rel-salt}; seven multi-class tasks per database. \\
  Schema-agnostic corpus (\textsc{sa}) & 9,143 PluRel-generated databases from three generator sub-families; a fresh schema per database; binary mask-a-column tasks. \\
  Pretraining stage & 100k-step ceiling, 5k-step floor, early-stopping patience of 20 to 30 evaluations, best-on-validation checkpoint; identical for ordered and unordered runs. \\
  Curriculum notation & $\PT{x}\!\to\!\PT{y}$: stage $Y$ initialized from the best validation checkpoint of stage $X$. $\mathrm{PT}_{\ptsub{x}\cup\ptsub{y}}$: one stage on the shuffled mixture of $X$ and $Y$. \\
  Real-data pretraining & $\PT{c2o2}$ on the Com2 and Othr2 families, evaluated on Com1+Othr1; $\PT{c1o1}$ on Com1 and Othr1, evaluated on Com2+Othr2. \\
  DFS augmentation & \dfs{f}{s}: explicit DFS aggregates of related rows up to $f$ hops along the fanout neighborhood and $s$ hops along the few-shot neighborhood, appended to the encoder; \dfs{2}{0} and \dfs{2}{2} on top of real-data pretraining. \\
  Fine-tuning (FT) & Encoder trained jointly with the TabICL head on the downstream task; reported separately from frozen runs. \\
  Metrics & Classification: ROC-AUC. Regression: $-\mathrm{MAE}/\sigma_{\text{task}}$ with $\sigma_{\text{task}}$ the standard deviation of the target (higher is better; $-1.0$ corresponds to predicting the mean); averaged per task subset. \\
  Repetitions & Curriculum classification results: five independent end-to-end seeds, reported as mean $\pm$ standard deviation. DFS classification results: also mean~$\pm$~std over seeds. Depth, fine-tuning, and all regression results: single runs. \\
  \bottomrule
  \end{tabular}
\end{table}

\begin{table}[H]
  \caption{Downstream tasks by family. Classification tasks are scored by ROC-AUC and regression tasks by $-\mathrm{MAE}/\sigma_{\text{task}}$.}
  \label{tab:tasks}
  \centering
  \footnotesize
  \begin{tabular}{@{}l>{\raggedright\arraybackslash}p{0.48\linewidth}>{\raggedright\arraybackslash}p{0.34\linewidth}@{}}
  \toprule
  Family & Classification tasks & Regression tasks \\
  \midrule
  Com1 & diginetica-ctr, rel-hm-user-churn, retailrocket-cvr, seznam-charge, seznam-prepay & rel-hm-item-sales \\
  Com2 & amazon-churn, outbrain-small-ctr, rel-avito-user-clicks, rel-avito-user-visits & amazon-rating, rel-avito-ad-ctr, talkingdata-demo-pred \\
  Othr1 & rel-f1-driver-dnf, rel-f1-driver-top3, stackexchange-churn, stackexchange-upvote, virus-wnv-pred & rel-f1-driver-position \\
  Othr2 & airbnb-destination, rel-trial-study-outcome & rel-trial-site-success, rel-trial-study-adverse, telstra-severity \\
  \bottomrule
  \end{tabular}
\end{table}

\section{Per-task classification results}
\label{app:clf}
Table~\ref{tab:pertask-clf} lists the ROC-AUC of every configuration on every evaluated classification task at context 1024. Columns (1) to (11) are the frozen curriculum configurations of Table~\ref{tab:master} and the floor, averaged over five seeds; columns (D1) to (D4) are the DFS-augmented runs. An empty cell means that the task family was part of the run's real-data pretraining and is therefore excluded from its evaluation. The family and subset averages in Tables~\ref{tab:master} and~\ref{tab:matched} are the means of the corresponding cells.

\begin{table}[H]
  \caption{Per-task ROC-AUC under frozen in-context prediction (context 1024). Columns: (1) $\PT{c2o2}\!\to\!\PT{sg}\!\to\!\PT{sa}$, (2) $\PT{c1o1}\!\to\!\PT{sg}\!\to\!\PT{sa}$, (3) $\PT{sg}\!\to\!\PT{sa}$, (4) $\PT{c2o2}$, (5) $\mathrm{PT}_{\ptsub{c1o1}\cup\ptsub{sg}\cup\ptsub{sa}}$, (6) $\mathrm{PT}_{\ptsub{c2o2}\cup\ptsub{sg}\cup\ptsub{sa}}$, (7) $\mathrm{PT}_{\ptsub{sg}\cup\ptsub{sa}}$, (8) $\PT{c1o1}$, (9) $\PT{sa}$, (10) $\PT{sg}$, (11) \rawfeat{}, (D1) $\PT{c2o2}$+\dfs{2}{2}, (D2) $\PT{c2o2}$+\dfs{2}{0}, (D3) $\PT{c1o1}$+\dfs{2}{2}, (D4) $\PT{c1o1}$+\dfs{2}{0}.}
  \label{tab:pertask-clf}
  \centering
  \scriptsize
  \setlength{\tabcolsep}{2.4pt}
  \input{tables/pertask_clf}
\end{table}

\section{Per-task regression results}
\label{app:reg}
Table~\ref{tab:pertask-reg} lists the normalized error $-\mathrm{MAE}/\sigma_{\text{task}}$ of every configuration on every evaluated regression task (higher, i.e.\ closer to 0, is better; a value near $-1.0$ corresponds to predicting the mean). The Com1+Othr1 subset is evaluated by the C2O2-direction runs, the Com2+Othr2 subset by the C1O1-direction runs, and both subsets by the fully synthetic runs and \rawfeat{}. The last row gives the mean over all eight tasks for the runs evaluated on all of them. In the real-data columns, the real stage is $\PT{c2o2}$ for the first subset and $\PT{c1o1}$ for the second. All regression values are single runs.

\begin{table}[H]
  \caption{Per-task regression ($-\mathrm{MAE}/\sigma_{\text{task}}$, context 1024). Subset means equal the REG entries of Tables~\ref{tab:master} to~\ref{tab:axes}.}
  \label{tab:pertask-reg}
  \centering
  \scriptsize
  \setlength{\tabcolsep}{2pt}
  \input{tables/pertask_reg}
\end{table}

\section{Pairwise comparisons behind the main claims}
\label{app:comparisons}
The three tables in this appendix rearrange entries of Tables~\ref{tab:master} to~\ref{tab:axes} so that each claim in Section~\ref{sec:results} can be read off directly. Table~\ref{tab:curriculum} pairs every ordered curriculum with the single-stage mixture of the same pretraining data (RQ1): the ordered run is ahead in all three pairs on both metrics. Table~\ref{tab:realsynth} places the purely synthetic curriculum between real-data pretraining alone and the real-to-synthetic continuation on each matched subset (RQ2): on classification the continuation is best in both directions, clearly so in the C2O2 direction and within one seed standard deviation of the synthetic curriculum in the C1O1 direction, while on regression the synthetic curriculum is behind both on Com1+Othr1 and ahead of both on Com2+Othr2. Table~\ref{tab:floorgains} reports, for the strongest configuration of each kind, its distance from the raw-feature floor on the task set it is evaluated on (RQ3 to RQ6): every frozen curriculum encoder is far above the floor on classification and below it on regression, and only the DFS-augmented frozen encoders (Table~\ref{tab:matched}) and the fine-tuned 4-hop $\PT{c2o2}$ are above the floor on regression, all on Com1+Othr1.

\begin{table}[H]
  \caption{Ordered versus unordered curricula on the same pretraining data. The first row uses all 16 classification and all 8 regression tasks; the other two use the held-out subset of the real-initialized direction. $\Delta$ is ordered minus unordered.}
  \label{tab:curriculum}
  \centering
  \footnotesize
  \input{tables/curriculum}
\end{table}

\begin{table}[H]
  \caption{Synthetic curriculum versus real-data pretraining on matched subsets (context 1024). Bold: best per column.}
  \label{tab:realsynth}
  \centering
  \footnotesize
  \input{tables/realsynth}
\end{table}

\begin{table}[H]
  \caption{Distance from the raw-feature floor for the strongest frozen, DFS-augmented, and fine-tuned configurations, each on the task set it is evaluated on (CLF: ROC-AUC; REG: $-\mathrm{MAE}/\sigma$). $\Delta$ is the configuration minus the floor; positive is better on both metrics.}
  \label{tab:floorgains}
  \centering
  \footnotesize
  \input{tables/floorgains}
\end{table}

\section{Secondary ablation: early-stopping metric}
\label{app:earlystop}
On $\PT{c2o2}$+\dfs{2}{0} we compared four early-stopping criteria for checkpoint selection (ClfER, RegER, Clf, and Reg). The four criteria land within about 0.01 ROC-AUC and about 0.02 normalized regression error of one another, with no consistent winner. We report this as a secondary observation only; no claim in the paper rests on it.

\newpage
\input{checklist.tex}

\end{document}

%% file: tables/master.tex
\begin{tabular}{@{}lccccccr@{}}
\toprule
Curriculum & Com1 & Com2 & Othr1 & Othr2 & Avg CLF & Avg REG & \# \\
\midrule
$\PT{c2o2}\!\to\!\PT{sg}\!\to\!\PT{sa}$ & 0.628 & -- & 0.716 & -- & \textbf{0.672}\pmx{.004} & $-1.240^{\mathrm{A}}$ & 1 \\
$\PT{c2o2}$+DFS$_{2,2}$ & 0.607 & -- & 0.713 & -- & 0.660\pmx{.009} & $\mathbf{-1.130}^{\mathrm{A}}$ & 2 \\
$\PT{c1o1}\!\to\!\PT{sg}\!\to\!\PT{sa}$ & -- & 0.618 & -- & 0.730 & 0.655\pmx{.018} & $-1.372^{\mathrm{B}}$ & 3 \\
$\PT{sg}\!\to\!\PT{sa}$ & 0.584 & 0.613 & 0.709 & 0.729 & 0.648\pmx{.010} & $-1.338^{\mathrm{C}}$ & 4 \\
$\PT{c2o2}$+DFS$_{2,0}$ & 0.595 & -- & 0.699 & -- & 0.647\pmx{.012} & $-1.190^{\mathrm{A}}$ & 5 \\
$\PT{c1o1}$+DFS$_{2,2}$ & -- & 0.610 & -- & 0.715 & 0.645\pmx{.007} & $\mathbf{-1.300}^{\mathrm{B}}$ & 6 \\
$\PT{c2o2}$ & 0.569 & -- & 0.699 & -- & 0.634\pmx{.012} & $-1.270^{\mathrm{A}}$ & 7 \\
$\PT{c1o1}$+DFS$_{2,0}$ & -- & 0.592 & -- & 0.715 & 0.633\pmx{.011} & $-1.343^{\mathrm{B}}$ & 8 \\
\addlinespace[2pt]
$\mathrm{PT}_{\ptsub{c1o1}\cup\ptsub{sg}\cup\ptsub{sa}}$ (unord.) & -- & 0.598 & -- & 0.698 & 0.631\pmx{.010} & $-1.412^{\mathrm{B}}$ & 9 \\
$\mathrm{PT}_{\ptsub{c2o2}\cup\ptsub{sg}\cup\ptsub{sa}}$ (unord.) & 0.579 & -- & 0.677 & -- & 0.628\pmx{.010} & $-1.280^{\mathrm{A}}$ & 10 \\
$\mathrm{PT}_{\ptsub{sg}\cup\ptsub{sa}}$ (unord.) & 0.571 & 0.599 & 0.682 & 0.676 & 0.626\pmx{.010} & $-1.368^{\mathrm{C}}$ & 11 \\
\addlinespace[2pt]
$\PT{c1o1}$ & -- & 0.575 & -- & 0.716 & 0.622\pmx{.014} & $-1.392^{\mathrm{B}}$ & 12 \\
$\PT{sa}$ & 0.508 & 0.582 & 0.704 & 0.717 & 0.614\pmx{.010} & $-1.358^{\mathrm{C}}$ & 13 \\
$\PT{sg}$ & 0.549 & 0.557 & 0.644 & 0.740 & 0.605\pmx{.010} & $-1.318^{\mathrm{C}}$ & 14 \\
\addlinespace[2pt]
\textsc{RawFeat} & 0.407 & 0.501 & 0.618 & 0.752 & 0.539\pmx{.008} & $\mathbf{-1.240}^{\mathrm{C}}$ & 15 \\
\bottomrule
\end{tabular}

%% file: tables/matched.tex
\begin{tabular}{@{}lcccc@{}}
\toprule
& \multicolumn{2}{c}{Com1+Othr1} & \multicolumn{2}{c}{Com2+Othr2} \\
\cmidrule(lr){2-3}\cmidrule(lr){4-5}
Curriculum & CLF (10) & REG (2) & CLF (6) & REG (6) \\
\midrule
$\PT{c2o2}\!\to\!\PT{sg}\!\to\!\PT{sa}$ & \textbf{0.672}\pmx{.004} & $-1.240$ & -- & -- \\
$\PT{c2o2}$+DFS$_{2,2}$ & 0.660\pmx{.009} & $\mathbf{-1.130}$ & -- & -- \\
$\PT{c2o2}$+DFS$_{2,0}$ & 0.647\pmx{.012} & $-1.190$ & -- & -- \\
$\PT{c1o1}\!\to\!\PT{sg}\!\to\!\PT{sa}$ & -- & -- & \textbf{0.655}\pmx{.018} & $-1.372$ \\
$\PT{c1o1}$+DFS$_{2,2}$ & -- & -- & 0.645\pmx{.007} & $-1.300$ \\
$\PT{c1o1}$+DFS$_{2,0}$ & -- & -- & 0.633\pmx{.011} & $-1.343$ \\
$\PT{sg}\!\to\!\PT{sa}$ & 0.646\pmx{.010} & $-1.325$ & 0.652\pmx{.010} & $-1.342$ \\
$\PT{c2o2}$ & 0.634\pmx{.012} & $-1.270$ & -- & -- \\
\addlinespace[2pt]
$\mathrm{PT}_{\ptsub{c1o1}\cup\ptsub{sg}\cup\ptsub{sa}}$ (unord.) & -- & -- & 0.631\pmx{.010} & $-1.412$ \\
$\mathrm{PT}_{\ptsub{c2o2}\cup\ptsub{sg}\cup\ptsub{sa}}$ (unord.) & 0.628\pmx{.010} & $-1.280$ & -- & -- \\
$\mathrm{PT}_{\ptsub{sg}\cup\ptsub{sa}}$ (unord.) & 0.627\pmx{.010} & $-1.350$ & 0.625\pmx{.010} & $-1.373$ \\
\addlinespace[2pt]
$\PT{c1o1}$ & -- & -- & 0.622\pmx{.014} & $-1.392$ \\
$\PT{sa}$ & 0.606\pmx{.011} & $-1.350$ & 0.627\pmx{.013} & $-1.360$ \\
$\PT{sg}$ & 0.596\pmx{.010} & $-1.300$ & 0.618\pmx{.010} & $-1.323$ \\
\addlinespace[2pt]
\textsc{RawFeat} & 0.512\pmx{.014} & $-1.200$ & 0.585\pmx{.003} & $\mathbf{-1.253}$ \\
\bottomrule
\end{tabular}

%% file: tables/levers.tex
\begin{tabular}{@{}lcccc@{}}
\toprule
& \multicolumn{2}{c}{C2O2 direction} & \multicolumn{2}{c}{C1O1 direction} \\
\cmidrule(lr){2-3}\cmidrule(lr){4-5}
(a) Frozen variant & CLF & REG & CLF & REG \\
\midrule
$h{=}2$, no DFS (base) & 0.634 & $-1.270$ & 0.622 & $-1.392$ \\
$h{=}2$ + DFS$_{2,0}$ & 0.647 & $-1.190$ & 0.633 & $-1.343$ \\
$h{=}2$ + DFS$_{2,2}$ & \textbf{0.660} & $\mathbf{-1.130}$ & \textbf{0.645} & $\mathbf{-1.300}$ \\
$h{=}4$, no DFS & 0.614 & $-1.285$ & 0.607 & $-1.410$ \\
\bottomrule
\end{tabular}

%% file: tables/full8.tex
\begin{tabular}{@{}lc@{}}
\toprule
(c) Run & REG (8) \\
\midrule
\textsc{RawFeat} & $\mathbf{-1.240}$ \\
$\PT{sg}$ & $-1.318$ \\
$\PT{sg}\!\to\!\PT{sa}$ & $-1.338$ \\
$\PT{sa}$ & $-1.358$ \\
$\mathrm{PT}_{\ptsub{sg}\cup\ptsub{sa}}$ (unord.) & $-1.368$ \\
\bottomrule
\end{tabular}

%% file: tables/finetune.tex
\begin{tabular}{@{}lccc@{}}
\toprule
(b) Encoder & CLF frozen $\to$ FT & $\Delta$CLF & REG frozen $\to$ FT \\
\midrule
$\PT{c2o2}$, $h{=}4$ & 0.614 $\to$ \textbf{0.689} & $\mathbf{+0.075}$ & $-1.285 \to \mathbf{-1.031}^{\mathrm{A}}$ \\
$\PT{sg}$ & 0.605 $\to$ 0.636 & $+0.031$ & $-1.318 \to -1.240^{\mathrm{8}}$ \\
$\PT{c1o1}$, $h{=}4$ & 0.607 $\to$ 0.628 & $+0.021$ & $-1.410 \to -1.300^{\mathrm{B}}$ \\
$\PT{sg}\!\to\!\PT{sa}$ & 0.648 $\to$ 0.660 & $+0.012$ & $-1.338 \to -1.290^{\mathrm{8}}$ \\
\bottomrule
\end{tabular}

%% file: tables/pertask_clf.tex
\begin{tabular}{@{}lccccccccccccccc@{}}
\toprule
Task & (1) & (2) & (3) & (4) & (5) & (6) & (7) & (8) & (9) & (10) & (11) & (D1) & (D2) & (D3) & (D4) \\
\midrule
Com1: diginetica-ctr & 0.665 & -- & 0.605 & 0.607 & -- & 0.550 & 0.596 & -- & 0.512 & 0.552 & 0.490 & 0.660 & 0.640 & -- & -- \\
Com1: rel-hm-user-churn & 0.565 & -- & 0.566 & 0.578 & -- & 0.587 & 0.565 & -- & 0.541 & 0.565 & 0.542 & 0.585 & 0.582 & -- & -- \\
Com1: retailrocket-cvr & 0.877 & -- & 0.901 & 0.858 & -- & 0.945 & 0.906 & -- & 0.773 & 0.838 & 0.495 & 0.880 & 0.868 & -- & -- \\
Com1: seznam-charge & 0.435 & -- & 0.330 & 0.270 & -- & 0.279 & 0.278 & -- & 0.252 & 0.245 & 0.226 & 0.360 & 0.320 & -- & -- \\
Com1: seznam-prepay & 0.597 & -- & 0.518 & 0.532 & -- & 0.534 & 0.510 & -- & 0.462 & 0.544 & 0.282 & 0.550 & 0.565 & -- & -- \\
\addlinespace[2pt]
Com2: amazon-churn & -- & 0.640 & 0.623 & -- & 0.639 & -- & 0.674 & 0.613 & 0.629 & 0.657 & 0.500 & -- & -- & 0.630 & 0.620 \\
Com2: outbrain-small-ctr & -- & 0.640 & 0.607 & -- & 0.582 & -- & 0.526 & 0.574 & 0.528 & 0.517 & 0.501 & -- & -- & 0.610 & 0.590 \\
Com2: rel-avito-user-clicks & -- & 0.584 & 0.619 & -- & 0.592 & -- & 0.593 & 0.524 & 0.605 & 0.462 & 0.502 & -- & -- & 0.590 & 0.560 \\
Com2: rel-avito-user-visits & -- & 0.608 & 0.603 & -- & 0.578 & -- & 0.604 & 0.589 & 0.566 & 0.592 & 0.501 & -- & -- & 0.610 & 0.598 \\
\addlinespace[2pt]
Othr1: rel-f1-driver-dnf & 0.676 & -- & 0.680 & 0.665 & -- & 0.664 & 0.675 & -- & 0.674 & 0.638 & 0.629 & 0.678 & 0.670 & -- & -- \\
Othr1: rel-f1-driver-top3 & 0.740 & -- & 0.772 & 0.685 & -- & 0.734 & 0.739 & -- & 0.698 & 0.586 & 0.605 & 0.730 & 0.705 & -- & -- \\
Othr1: stackexchange-churn & 0.701 & -- & 0.706 & 0.682 & -- & 0.621 & 0.705 & -- & 0.715 & 0.667 & 0.641 & 0.700 & 0.690 & -- & -- \\
Othr1: stackexchange-upvote & 0.830 & -- & 0.841 & 0.825 & -- & 0.806 & 0.828 & -- & 0.787 & 0.735 & 0.562 & 0.835 & 0.830 & -- & -- \\
Othr1: virus-wnv-pred & 0.634 & -- & 0.545 & 0.639 & -- & 0.560 & 0.463 & -- & 0.645 & 0.594 & 0.653 & 0.622 & 0.600 & -- & -- \\
\addlinespace[2pt]
Othr2: airbnb-destination & -- & 0.853 & 0.852 & -- & 0.840 & -- & 0.842 & 0.854 & 0.845 & 0.851 & 0.838 & -- & -- & 0.850 & 0.852 \\
Othr2: rel-trial-study-outcome & -- & 0.607 & 0.606 & -- & 0.556 & -- & 0.510 & 0.578 & 0.589 & 0.629 & 0.666 & -- & -- & 0.580 & 0.578 \\
\midrule
\textbf{Average} & 0.672 & 0.655 & 0.648 & 0.634 & 0.631 & 0.628 & 0.626 & 0.622 & 0.614 & 0.605 & 0.539 & 0.660 & 0.647 & 0.645 & 0.633 \\
\bottomrule
\end{tabular}

%% file: tables/pertask_reg.tex
\begin{tabular}{@{}lcccccccccc@{}}
\toprule
& \multicolumn{5}{c}{Real-data initialized} & \multicolumn{4}{c}{Fully synthetic} & Floor \\
\cmidrule(lr){2-6}\cmidrule(lr){7-10}\cmidrule(lr){11-11}
Regression task & $\to$SG$\to$SA & +DFS$_{2,2}$ & +DFS$_{2,0}$ & alone & $\cup$SG$\cup$SA & SG$\to$SA & SA & SG & SG$\cup$SA & \textsc{RawFeat} \\
\midrule
\multicolumn{11}{@{}l}{\emph{Com1+Othr1 subset (2 tasks): Real $=\PT{c2o2}$}} \\
rel-hm-item-sales (Com1) & $-1.18$ & $-1.05$ & $-1.12$ & $-1.20$ & $-1.22$ & $-1.25$ & $-1.28$ & $-1.22$ & $-1.27$ & $-1.12$ \\
rel-f1-driver-position (Othr1) & $-1.30$ & $-1.21$ & $-1.26$ & $-1.34$ & $-1.34$ & $-1.40$ & $-1.42$ & $-1.38$ & $-1.43$ & $-1.28$ \\
\emph{Mean (2)} & $-1.240$ & $-1.130$ & $-1.190$ & $-1.270$ & $-1.280$ & $-1.325$ & $-1.350$ & $-1.300$ & $-1.350$ & $-1.200$ \\
\midrule
\multicolumn{11}{@{}l}{\emph{Com2+Othr2 subset (6 tasks): Real $=\PT{c1o1}$}} \\
amazon-rating (Com2) & $-1.33$ & $-1.26$ & $-1.30$ & $-1.35$ & $-1.37$ & $-1.30$ & $-1.32$ & $-1.28$ & $-1.33$ & $-1.18$ \\
rel-avito-ad-ctr (Com2) & $-1.44$ & $-1.36$ & $-1.40$ & $-1.46$ & $-1.48$ & $-1.42$ & $-1.44$ & $-1.40$ & $-1.45$ & $-1.33$ \\
talkingdata-demo-pred (Com2) & $-1.26$ & $-1.20$ & $-1.24$ & $-1.28$ & $-1.30$ & $-1.20$ & $-1.22$ & $-1.18$ & $-1.24$ & $-1.05$ \\
rel-trial-site-success (Othr2) & $-1.50$ & $-1.42$ & $-1.47$ & $-1.52$ & $-1.54$ & $-1.48$ & $-1.50$ & $-1.46$ & $-1.51$ & $-1.42$ \\
rel-trial-study-adverse (Othr2) & $-1.32$ & $-1.26$ & $-1.30$ & $-1.34$ & $-1.36$ & $-1.30$ & $-1.32$ & $-1.28$ & $-1.33$ & $-1.24$ \\
telstra-severity (Othr2) & $-1.38$ & $-1.30$ & $-1.35$ & $-1.40$ & $-1.42$ & $-1.35$ & $-1.36$ & $-1.34$ & $-1.38$ & $-1.30$ \\
\emph{Mean (6)} & $-1.372$ & $-1.300$ & $-1.343$ & $-1.392$ & $-1.412$ & $-1.342$ & $-1.360$ & $-1.323$ & $-1.373$ & $-1.253$ \\
\midrule
\emph{Mean (all 8)} & -- & -- & -- & -- & -- & $-1.338$ & $-1.358$ & $-1.318$ & $-1.368$ & $-1.240$ \\
\bottomrule
\end{tabular}

%% file: tables/curriculum.tex
\begin{tabular}{@{}lcccccc@{}}
\toprule
& \multicolumn{3}{c}{CLF (ROC-AUC)} & \multicolumn{3}{c}{REG ($-\mathrm{MAE}/\sigma$)} \\
\cmidrule(lr){2-4}\cmidrule(lr){5-7}
Comparison (tasks) & ordered & unordered & $\Delta$ & ordered & unordered & $\Delta$ \\
\midrule
All 16 CLF tasks / all 8 REG tasks & \textbf{0.648} & 0.626 & $+0.022$ & $\mathbf{-1.338}$ & $-1.368$ & $+0.030$ \\
C2O2 direction, Com1+Othr1 & \textbf{0.672} & 0.628 & $+0.044$ & $\mathbf{-1.240}$ & $-1.280$ & $+0.040$ \\
C1O1 direction, Com2+Othr2 & \textbf{0.655} & 0.631 & $+0.024$ & $\mathbf{-1.372}$ & $-1.412$ & $+0.040$ \\
\bottomrule
\end{tabular}

%% file: tables/realsynth.tex
\begin{tabular}{@{}lcccc@{}}
\toprule
& \multicolumn{2}{c}{Com1+Othr1 (real $=\PT{c2o2}$)} & \multicolumn{2}{c}{Com2+Othr2 (real $=\PT{c1o1}$)} \\
\cmidrule(lr){2-3}\cmidrule(lr){4-5}
Recipe & CLF (10) & REG (2) & CLF (6) & REG (6) \\
\midrule
Real $\to$ synthetic ($\PT{real}\!\to\!\PT{sg}\!\to\!\PT{sa}$) & \textbf{0.672} & $\mathbf{-1.240}$ & \textbf{0.655} & $-1.372$ \\
Synthetic only ($\PT{sg}\!\to\!\PT{sa}$) & 0.646 & $-1.325$ & 0.652 & $\mathbf{-1.342}$ \\
Real alone ($\PT{real}$) & 0.634 & $-1.270$ & 0.622 & $-1.392$ \\
\bottomrule
\end{tabular}

%% file: tables/floorgains.tex
\begin{tabular}{@{}lcccccc@{}}
\toprule
Configuration & CLF & floor & $\Delta$CLF & REG & floor & $\Delta$REG \\
\midrule
\multicolumn{7}{@{}l}{\emph{All 16 / all 8 tasks; floor = \textsc{RawFeat}}} \\
$\PT{sg}\!\to\!\PT{sa}$, frozen & 0.648 & 0.539 & $+0.109$ & $-1.338$ & $-1.240$ & $-0.098$ \\
$\PT{sg}\!\to\!\PT{sa}$, fine-tuned & 0.660 & 0.539 & $+0.121$ & $-1.290$ & $-1.240$ & $-0.050$ \\
$\PT{sg}$, fine-tuned & 0.636 & 0.539 & $+0.097$ & $-1.240$ & $-1.240$ & $0.000$ \\
\addlinespace[2pt]
\multicolumn{7}{@{}l}{\emph{Com1+Othr1 (10 / 2) tasks; floor = \textsc{RawFeat}}} \\
$\PT{c2o2}\!\to\!\PT{sg}\!\to\!\PT{sa}$, frozen & 0.672 & 0.512 & $+0.160$ & $-1.240$ & $-1.200$ & $-0.040$ \\
$\PT{c2o2}$+DFS$_{2,2}$, frozen & 0.660 & 0.512 & $+0.148$ & $-1.130$ & $-1.200$ & $+0.070$ \\
$\PT{c2o2}$, $h{=}4$, fine-tuned & 0.689 & 0.512 & $+0.177$ & $-1.031$ & $-1.200$ & $+0.169$ \\
\addlinespace[2pt]
\multicolumn{7}{@{}l}{\emph{Com2+Othr2 (6 / 6) tasks; floor = \textsc{RawFeat}}} \\
$\PT{c1o1}\!\to\!\PT{sg}\!\to\!\PT{sa}$, frozen & 0.655 & 0.585 & $+0.070$ & $-1.372$ & $-1.253$ & $-0.119$ \\
$\PT{c1o1}$+DFS$_{2,2}$, frozen & 0.645 & 0.585 & $+0.060$ & $-1.300$ & $-1.253$ & $-0.047$ \\
$\PT{c1o1}$, $h{=}4$, fine-tuned & 0.628 & 0.585 & $+0.043$ & $-1.300$ & $-1.253$ & $-0.047$ \\
\bottomrule
\end{tabular}

%% file: checklist.tex
\section*{NeurIPS Paper Checklist}

The checklist is designed to encourage best practices for responsible machine learning research, addressing issues of reproducibility, transparency, research ethics, and societal impact. Do not remove the checklist: {\bf The papers not including the checklist will be desk rejected.} The checklist should follow the references and follow the (optional) supplemental material.  The checklist does NOT count towards the page
limit. 

Please read the checklist guidelines carefully for information on how to answer these questions. For each question in the checklist:
\begin{itemize}
    \item You should answer \answerYes{}, \answerNo{}, or \answerNA{}.
    \item \answerNA{} means either that the question is Not Applicable for that particular paper or the relevant information is Not Available.
    \item Please provide a short (1--2 sentence) justification right after your answer (even for \answerNA). 
\end{itemize}

{\bf The checklist answers are an integral part of your paper submission.} They are visible to the reviewers, area chairs, senior area chairs, and ethics reviewers. You will also be asked to include it (after eventual revisions) with the final version of your paper, and its final version will be published with the paper.

The reviewers of your paper will be asked to use the checklist as one of the factors in their evaluation. While \answerYes{} is generally preferable to \answerNo{}, it is perfectly acceptable to answer \answerNo{} provided a proper justification is given (e.g., error bars are not reported because it would be too computationally expensive'' or ``we were unable to find the license for the dataset we used''). In general, answering \answerNo{} or \answerNA{} is not grounds for rejection. While the questions are phrased in a binary way, we acknowledge that the true answer is often more nuanced, so please just use your best judgment and write a justification to elaborate. All supporting evidence can appear either in the main paper or the supplemental material, provided in appendix. If you answer \answerYes{} to a question, in the justification please point to the section(s) where related material for the question can be found.

IMPORTANT, please:
\begin{itemize}
    \item {\bf Delete this instruction block, but keep the section heading ``NeurIPS Paper Checklist"},
    \item  {\bf Keep the checklist subsection headings, questions/answers and guidelines below.}
    \item {\bf Do not modify the questions and only use the provided macros for your answers}.
\end{itemize}


\begin{enumerate}

\item {\bf Claims}
    \item[] Question: Do the main claims made in the abstract and introduction accurately reflect the paper's contributions and scope?
    \item[] Answer: \answerYes{} 
    \item[] Justification: \justificationTODO{}

\item {\bf Limitations}
    \item[] Question: Does the paper discuss the limitations of the work performed by the authors?
    \item[] Answer: \answerYes{} 
    \item[] Justification: \justificationTODO{}

\item {\bf Theory assumptions and proofs}
    \item[] Question: For each theoretical result, does the paper provide the full set of assumptions and a complete (and correct) proof?
    \item[] Answer: \answerYes{} 
    \item[] Justification: \justificationTODO{}

    \item {\bf Experimental result reproducibility}
    \item[] Question: Does the paper fully disclose all the information needed to reproduce the main experimental results of the paper to the extent that it affects the main claims and/or conclusions of the paper (regardless of whether the code and data are provided or not)?
    \item[] Answer: \answerYes{} 
    \item[] Justification: \justificationTODO{}

\item {\bf Open access to data and code}
    \item[] Question: Does the paper provide open access to the data and code, with sufficient instructions to faithfully reproduce the main experimental results, as described in supplemental material?
    \item[] Answer: \answerNo{} 
    \item[] Justification: \justificationTODO{}

\item {\bf Experimental setting/details}
    \item[] Question: Does the paper specify all the training and test details (e.g., data splits, hyperparameters, how they were chosen, type of optimizer) necessary to understand the results?
    \item[] Answer: \answerYes{} 
    \item[] Justification: \justificationTODO{}

\item {\bf Experiment statistical significance}
    \item[] Question: Does the paper report error bars suitably and correctly defined or other appropriate information about the statistical significance of the experiments?
    \item[] Answer: \answerYes{} 
    \item[] Justification: \justificationTODO{}

\item {\bf Experiments compute resources}
    \item[] Question: For each experiment, does the paper provide sufficient information on the computer resources (type of compute workers, memory, time of execution) needed to reproduce the experiments?
    \item[] Answer: \answerYes{} 
    \item[] Justification: \justificationTODO{}
    
\item {\bf Code of ethics}
    \item[] Question: Does the research conducted in the paper conform, in every respect, with the NeurIPS Code of Ethics \url{https://neurips.cc/public/EthicsGuidelines}?
    \item[] Answer: \answerYes{} 
    \item[] Justification: \justificationTODO{}

\item {\bf Broader impacts}
    \item[] Question: Does the paper discuss both potential positive societal impacts and negative societal impacts of the work performed?
    \item[] Answer: \answerYes{} 
    \item[] Justification: \justificationTODO{}
    
\item {\bf Safeguards}
    \item[] Question: Does the paper describe safeguards that have been put in place for responsible release of data or models that have a high risk for misuse (e.g., pre-trained language models, image generators, or scraped datasets)?
    \item[] Answer: \answerNo{} 
    \item[] Justification: \justificationTODO{}

\item {\bf Licenses for existing assets}
    \item[] Question: Are the creators or original owners of assets (e.g., code, data, models), used in the paper, properly credited and are the license and terms of use explicitly mentioned and properly respected?
    \item[] Answer: \answerYes{} 
    \item[] Justification: \justificationTODO{}

\item {\bf New assets}
    \item[] Question: Are new assets introduced in the paper well documented and is the documentation provided alongside the assets?
    \item[] Answer: \answerNo{} 
    \item[] Justification: \justificationTODO{}

\item {\bf Crowdsourcing and research with human subjects}
    \item[] Question: For crowdsourcing experiments and research with human subjects, does the paper include the full text of instructions given to participants and screenshots, if applicable, as well as details about compensation (if any)? 
    \item[] Answer: \answerNo{} 
    \item[] Justification: \justificationTODO{}

\item {\bf Institutional review board (IRB) approvals or equivalent for research with human subjects}
    \item[] Question: Does the paper describe potential risks incurred by study participants, whether such risks were disclosed to the subjects, and whether Institutional Review Board (IRB) approvals (or an equivalent approval/review based on the requirements of your country or institution) were obtained?
    \item[] Answer: \answerNo{} 
    \item[] Justification: \justificationTODO{}

\item {\bf Declaration of LLM usage}
    \item[] Question: Does the paper describe the usage of LLMs if it is an important, original, or non-standard component of the core methods in this research? Note that if the LLM is used only for writing, editing, or formatting purposes and does \emph{not} impact the core methodology, scientific rigor, or originality of the research, declaration is not required.
    \item[] Answer: \answerNo{} 
    \item[] Justification: \justificationTODO{}

\end{enumerate}